\documentclass{article}
\usepackage{spconf,amsmath,graphicx}
\usepackage{spconf,amsmath,graphicx}
\usepackage{cite}
\usepackage{array}
\usepackage{xcolor}
\usepackage[utf8]{inputenc}
\usepackage{mathtools}
\usepackage{multirow}
\usepackage{algorithm}
\usepackage{algpseudocode}
\usepackage{subfigure}
\usepackage{setspace}
\usepackage{adjustbox}
\usepackage{caption}
\usepackage{lipsum}
\usepackage{bm,mathtools,amsmath,amsthm,amssymb,amsfonts,bbm}
\usepackage{breqn}
\usepackage{multicol}

\begin{document}
\title{
Robustness Evaluation of Machine Learning Models for Robot Arm Action Recognition in Noisy Environments}
%
\name{Elaheh Motamedi $^\dagger$,  Kian Behzad $^\dagger$, Rojin Zandi $^\dagger$, Hojjat Salehinejad $^{\ddagger\star}$, \textit{and Milad Siami $^\dagger$}
 \thanks{This material is based upon work supported in part at Northeastern University by grants ONR N00014-21-1-2431, NSF 2121121, the U.S. Department of Homeland Security under Grant Award Number 22STESE00001-01-00, and by the Army Research Laboratory under Cooperative Agreement Number W911NF-22-2-0001. The views and conclusions contained in this document are solely those of the authors and should not be interpreted as representing the official policies, either expressed or implied, of the U.S. Department of Homeland Security, the Army Research Office, or the U.S. Government.}
}
\address{$^\dagger$Department of Electrical \& Computer Engineering, Northeastern University, Boston, MA, USA\\
$^\ddagger$Kern Center for the Science of Health Care Delivery, Mayo Clinic, Rochester, MN, USA\\
$^\star$Department of Artificial Intelligence and Informatics, Mayo Clinic, Rochester, MN, USA}
%
%
%
\allowdisplaybreaks
\maketitle
%
%
\begin{abstract}
In the realm of robot action recognition, identifying distinct but spatially proximate arm movements using vision systems in noisy environments poses a significant challenge. 
This paper studies robot arm action recognition in noisy environments using machine learning techniques. Specifically, a vision system is used to track the robot's movements followed by a deep learning model to extract the arm's key points.  
Through a comparative analysis of machine learning methods, the effectiveness and robustness of this model are assessed in noisy environments. 
A case study was conducted using the Tic-Tac-Toe game in a 3-by-3 grid environment, where the focus is to accurately identify the actions of the arms in selecting specific locations within this constrained environment. 
Experimental results show that our approach can achieve precise key point detection and action classification despite the addition of noise and uncertainties to the dataset. 
\end{abstract}
\allowdisplaybreaks
\vspace{-0.2cm}
\begin{keywords}
Franka Emika robot arm, deep learning, key point extraction, noisy environment, robot arm action recognition. 
\end{keywords}
\allowdisplaybreaks

\section{Introduction}
Robotic systems are widely used in diverse sectors such as healthcare, manufacturing, and automation~\cite{lu2021super,lee2020camera}. Central to the performance and utility of these systems is the accurate determination of the robot's spatial orientation, commonly known as pose estimation, in an uncertain environment. Typically, pose estimation involves using regression models to detect the key points of a robot, which are the joints that make up its skeleton. Accurate pose estimation is crucial for various tasks, ranging from object manipulation to navigation and interaction with the environment \cite{klanvcar2014mobile,lu2021super,sun2022robust}.

With the advancement of machine learning, significant progress has been made in the field of robot pose detection\cite{heindl20193d,tian2023robot,lu2023markerless,zhou20193d}. As an example, convolutional neural networks (CNNs) have been widely developed to address the pose estimation problem of robot arms in ideal environment~\cite{gulde2018ropose, gulde2019ropose,  heindl2019learning,rodrigues2022framework,mivseikis2019two}. However, such solutions often overlook the challenges
of noisy and real-world environments, leading to models that excel theoretically but struggle in everyday scenarios. 

In this paper, a pretrained ResNet-50~\cite{he2016deep} is utilized as a standard tool for pose recognition. The output of this model is a time series of pose locations prone to noise. A CNN is proposed for robot arm action recognition from the noisy time series and its performance is compared with the state-of-the-art models such as transformers~\cite{ahmed2023transformers} and Rocket~\cite{dempster2020rocket}. The dataset for the experiments was collected using a Franka Emika~\cite{haddadin2022franka} robot arm. This study makes a significant contribution to the field of robot action recognition by introducing a model for robot arm action recognition with minimal error margins in noisy environments. The insights gained from this study have the potential to inform the development of more reliable robotic systems for various applications in noisy environments.

\begin{figure*}[t!]
    \begin{center}
         \includegraphics[width=.99\textwidth]
         {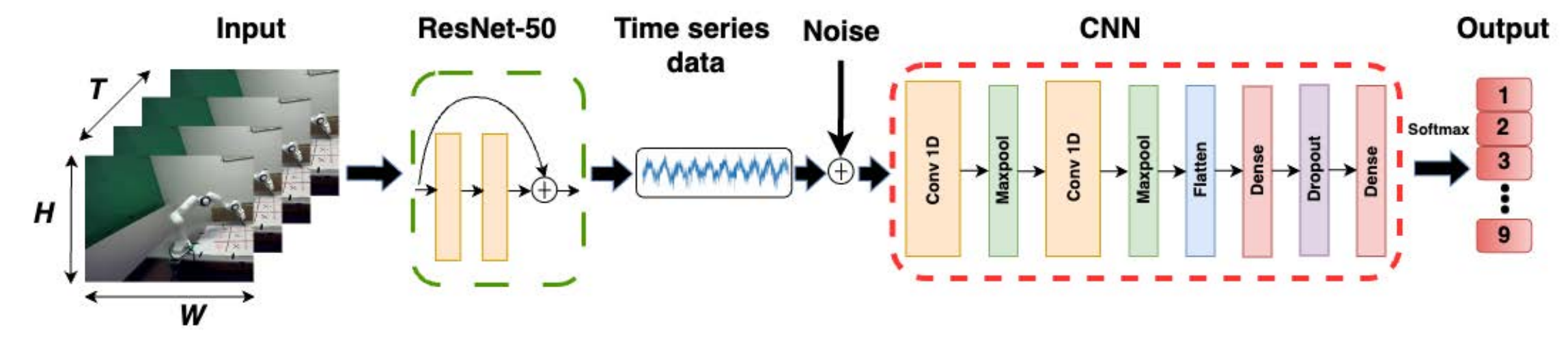}
    \end{center}
        \vspace{-0.3cm}
    \caption{Architecture of robot arm action recognition in noisy environments. Noise is applied after the estimation of the key points using ResNet-50 on the robot arm in the environment.}
    \label{fig:arch2}
\end{figure*}

\begin{figure}[t]
    \begin{center}
         \includegraphics[width=.32\textwidth]{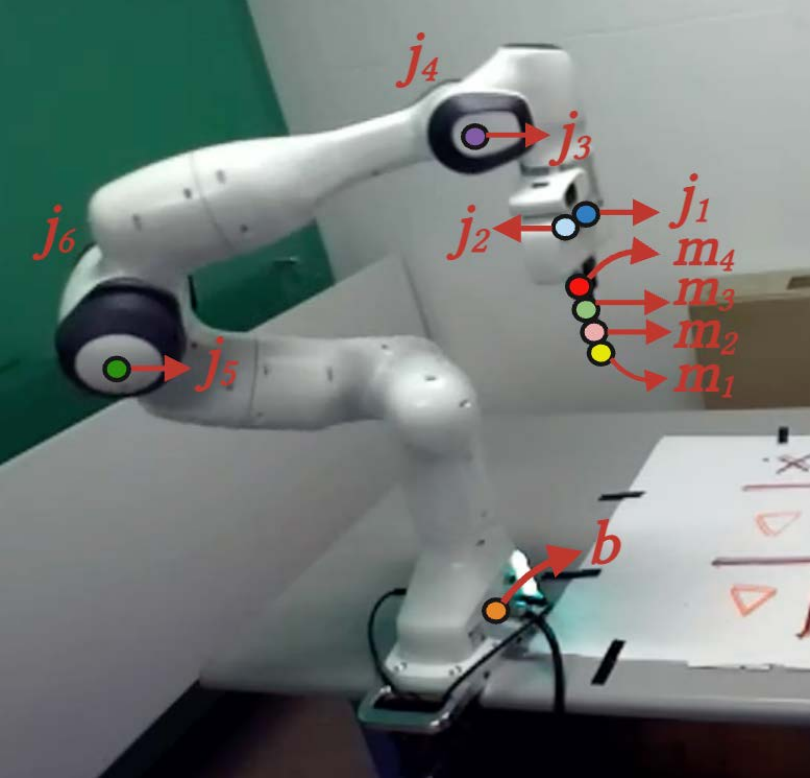}
    \end{center}
    \caption{Colored circles highlight detected key points, except for $ {j}_4$ and ${j}_6$, which are not visible in this frame due to visibility constraints.}
    \vspace{-0.5cm}
    \label{fig:DLC_labels}
\end{figure}

\section{Method}

The proposed model leverages a pre-trained ResNet-50~\cite{he2016deep} on ImageNet~\cite{5206848} for robot arm pose recognition followed by a CNN for robot arm action recognition in noisy environments, as depicted in Figure~\ref{fig:arch2}. Each step is described as follows.


\subsection{Robot Arm Pose Recognition}
To capture the movement of the robot arm, our initial step involves extracting precise arm poses using $J$ cameras, where each camera $j$ outputs a video stream $\mathbf{V}^{(j)}$ of length $T$ with frame size $H \times W$. Each video stream is processed by a pretrained ResNet-50 to estimate the robot arm poses in the X-Y plane over time as $\mathbf{A}^{(j)}\in\mathbb{R}^{K \times T}$, where $K$ is the number of key points $(m_1,\dots, m_4, j_1,\dots, j_6, b)$ as in Figure~\ref{fig:DLC_labels}.

A significant advantage of this approach is its effectiveness with a limited number of labeled frames, attributed to the pre-training of ResNet-50 on ImageNet~\cite{5206848}.
\subsection{Robot Arm Action Recognition}
The output of ResNet-50 model for camera $j$, denoted as $\mathbf{A}^{(j)}$, represents the evolving trajectory of key points on a robot arm over time. It is assumed that a noisy environment can alternate $\mathbf{A}^{(j)}$ and affect the robot arm action recognition. To tackle this challenge, a CNN is proposed as depicted in Figure~\ref{fig:arch2}. 

\begin{figure}[t]
    \begin{center}
         \includegraphics[scale=0.5]{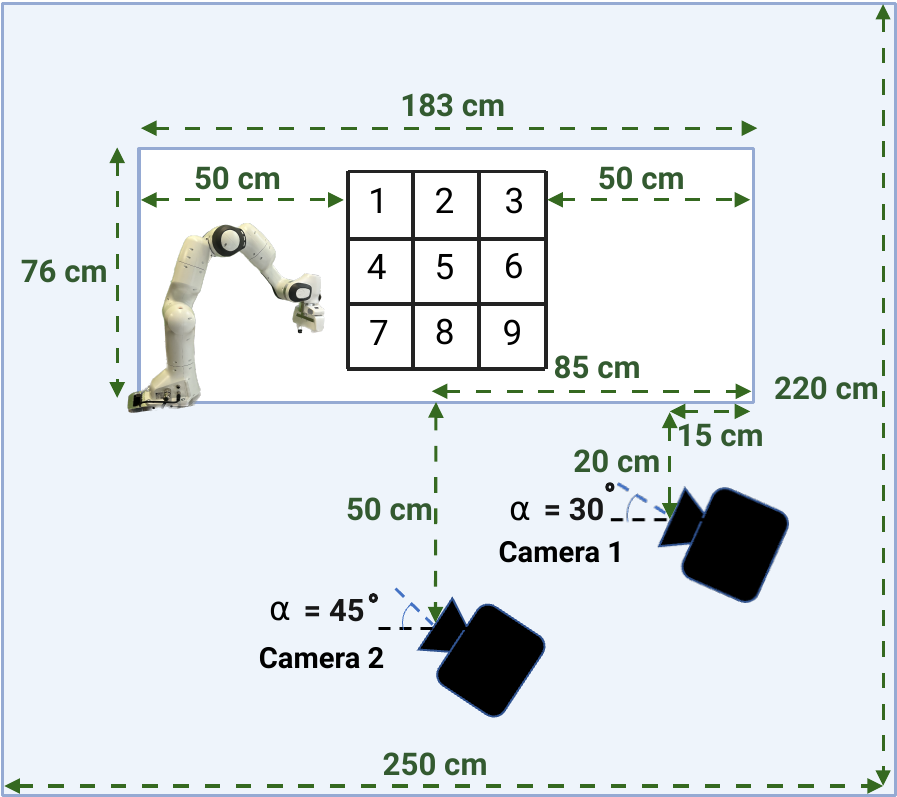}
    \end{center}
    \caption{Data collection using one robot arm and two cameras.}
    \label{fig:Labsettig}
\end{figure}

The proposed CNN model has a 1-D convolution layer with 32 filters where each filter is of size 3. After max-pooling with window size 2, another 1-D convolution layer is applied with 64 filters of kernel size 3. The resulted features after max-pooling with window size 2 are flattened and passed to two dense layers for action recognition. The number of possible target classes is 9, which represents different locations of the Tic-Tac-Toe game as illustrated in Figure~\ref{fig:Labsettig}.

\section{Experiments}
\subsection{Data Collection}
To evaluate our method, we utilized $J=2$  cameras positioned at varying locations within our laboratory. These cameras recorded the movements of the robot arm on Tic-Tac-Toe board, as depicted in Figure~\ref{fig:Labsettig}.

\begin{figure}[t!]
    \centering
      \begin{minipage}[b]{0.32\linewidth}
        \includegraphics[width=\linewidth]{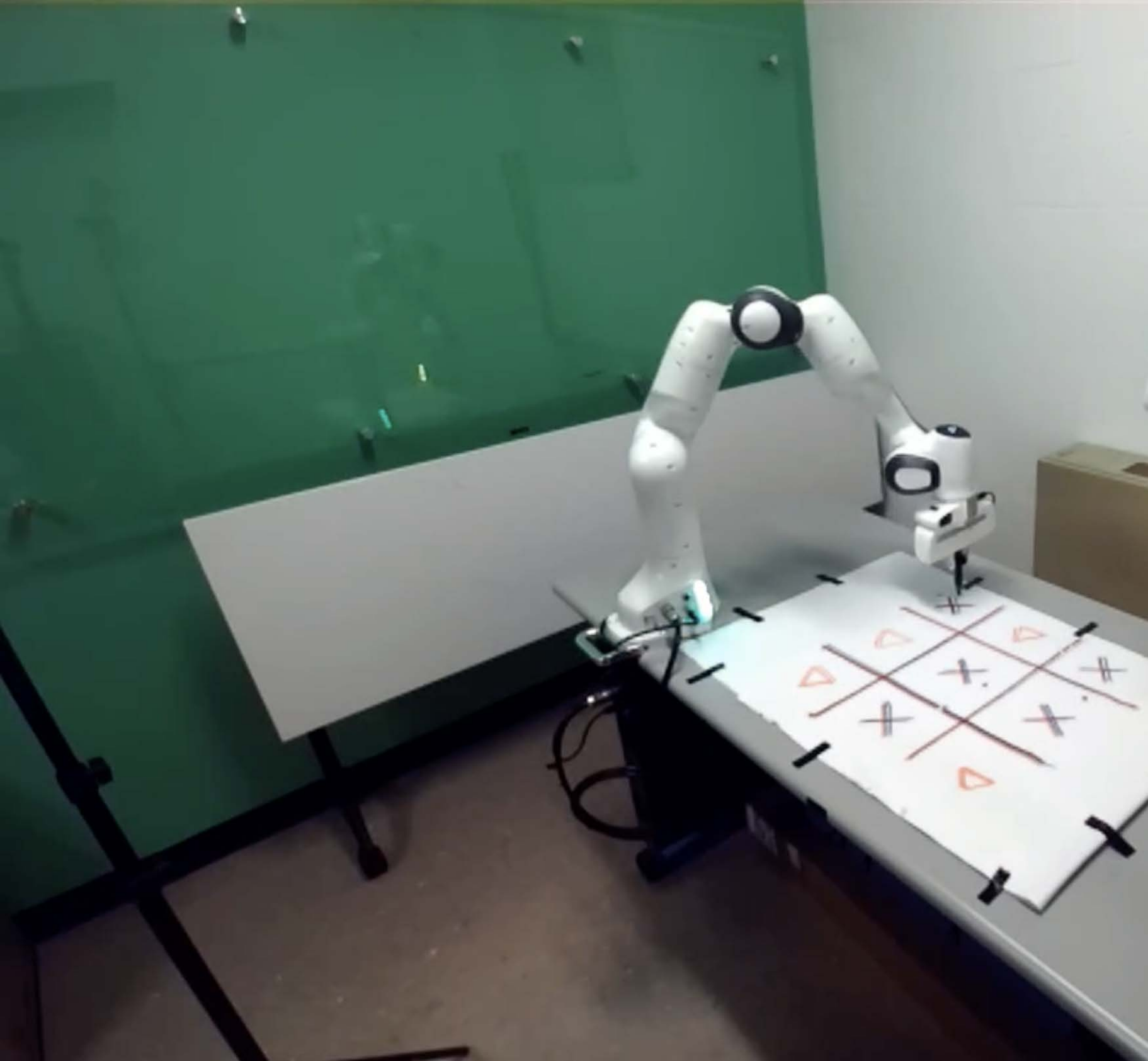}
      \end{minipage}
      \begin{minipage}[b]{0.32\linewidth}
        \includegraphics[width=\linewidth]{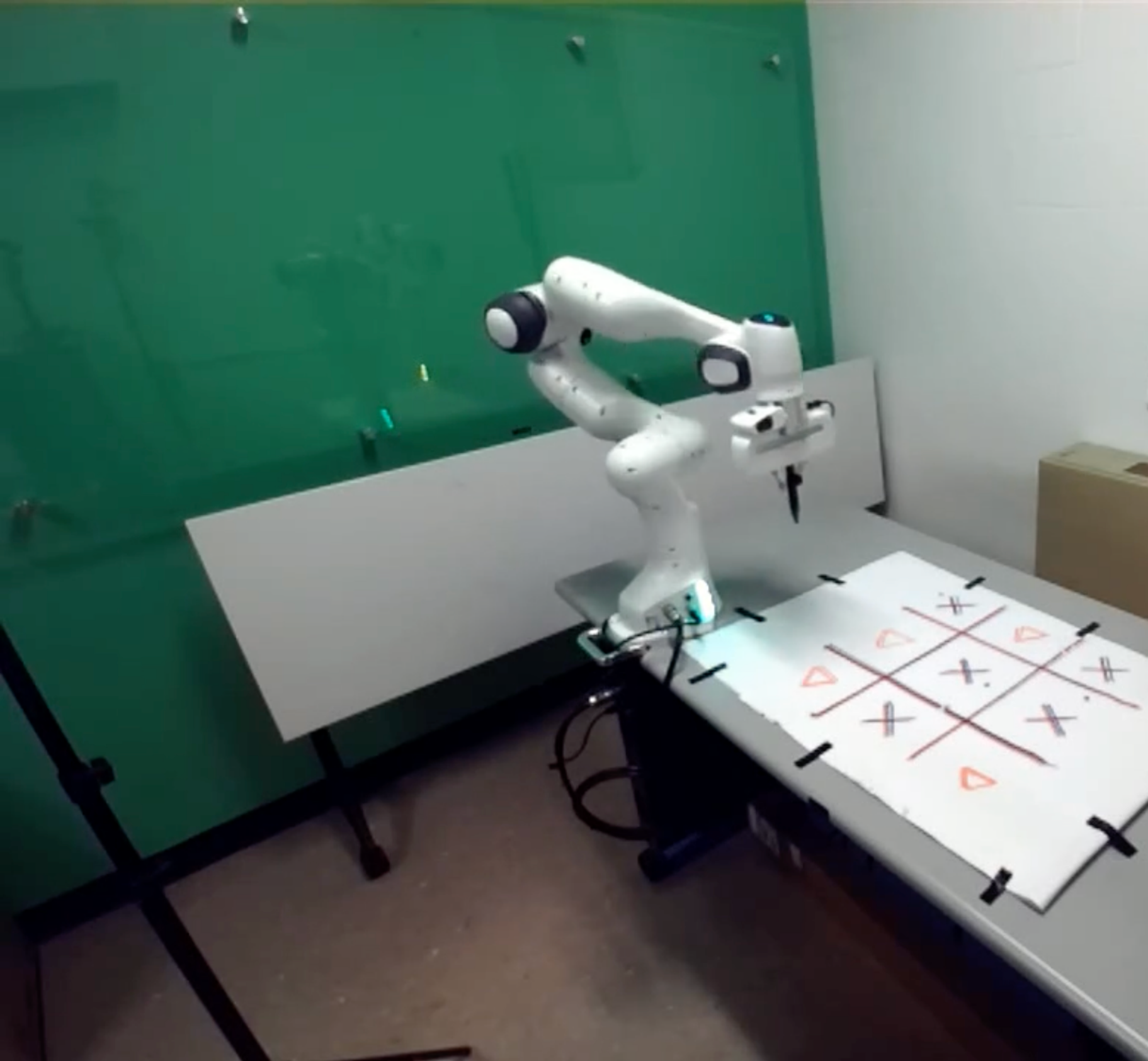}
      \end{minipage}
      \begin{minipage}[b]{0.32\linewidth}
        \includegraphics[width=\linewidth]{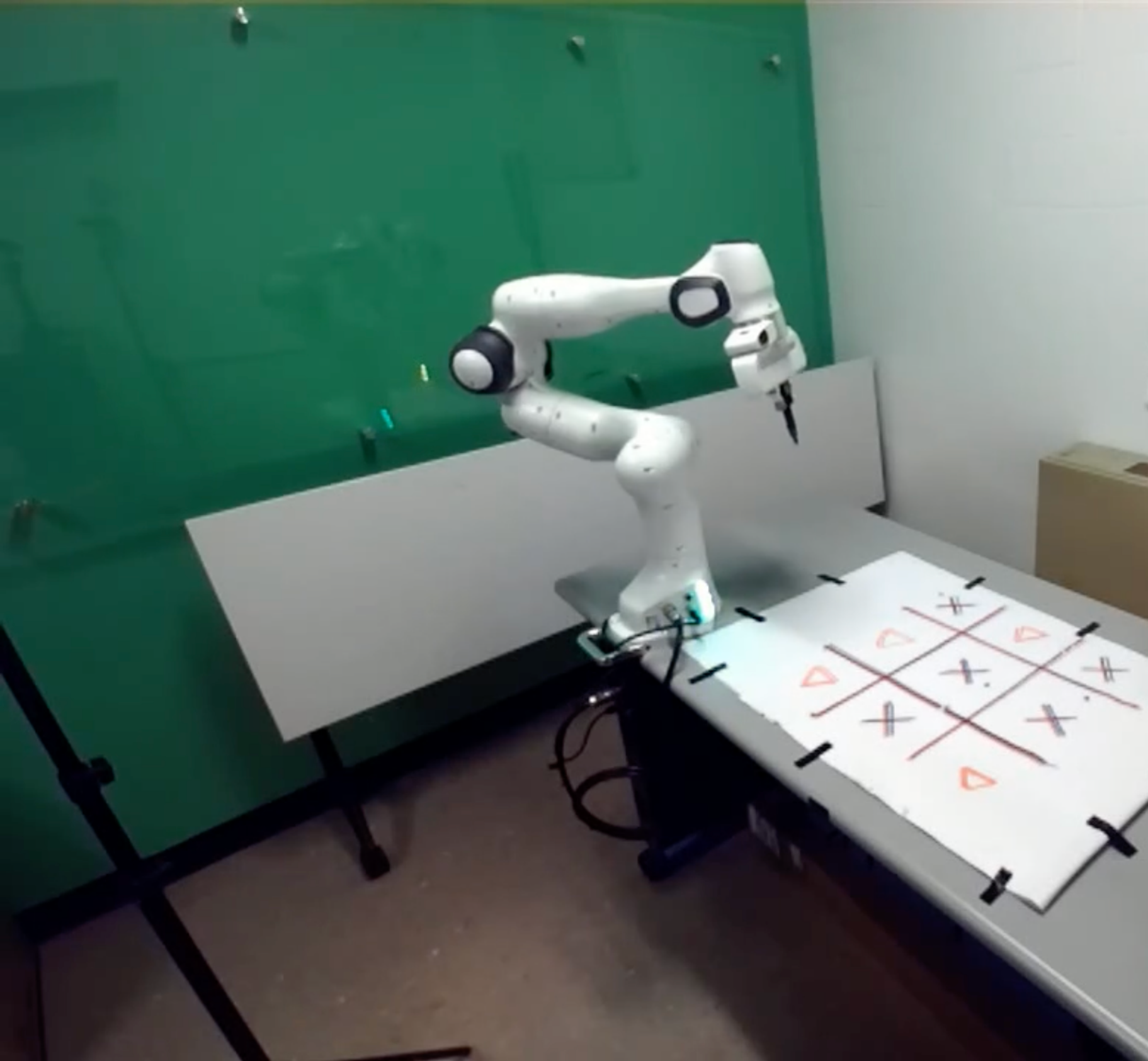}
      \end{minipage}\\
      \begin{minipage}[b]{0.32\linewidth}
        \includegraphics[width=\linewidth]{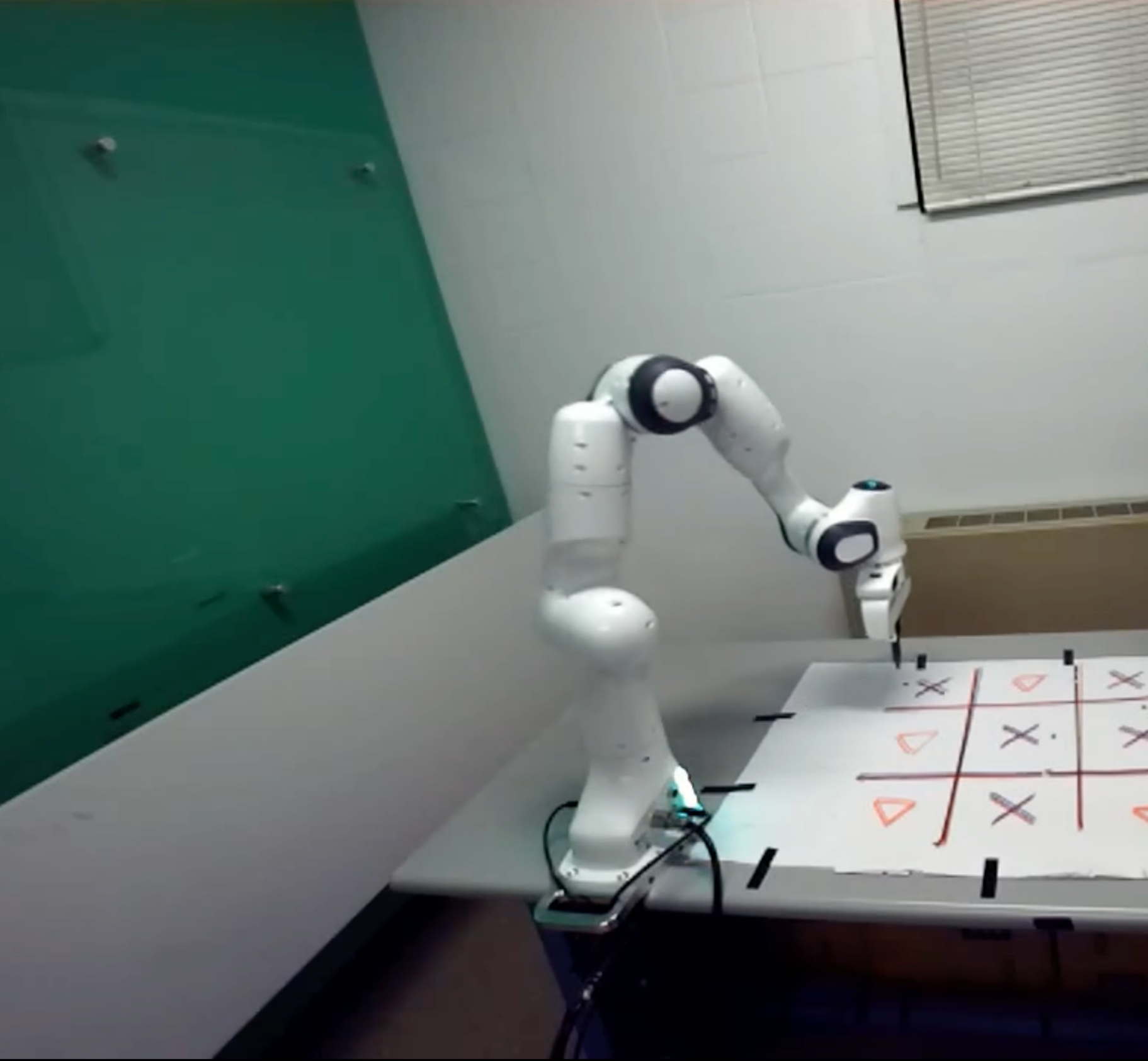}
      \end{minipage}
      \begin{minipage}[b]{0.32\linewidth}
        \includegraphics[width=\linewidth]{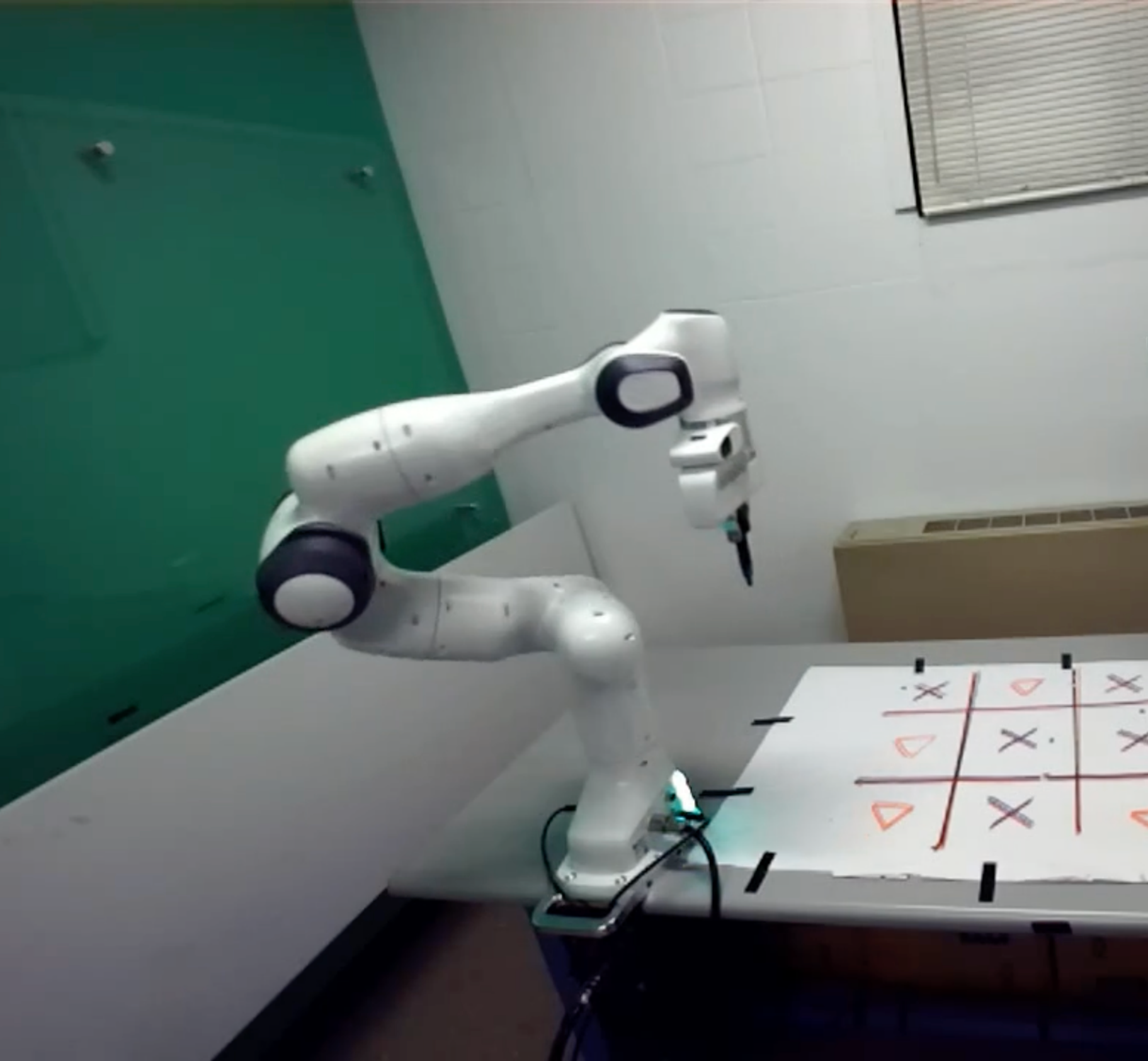}
      \end{minipage}
      \begin{minipage}[b]{0.32\linewidth}
        \includegraphics[width=\linewidth]{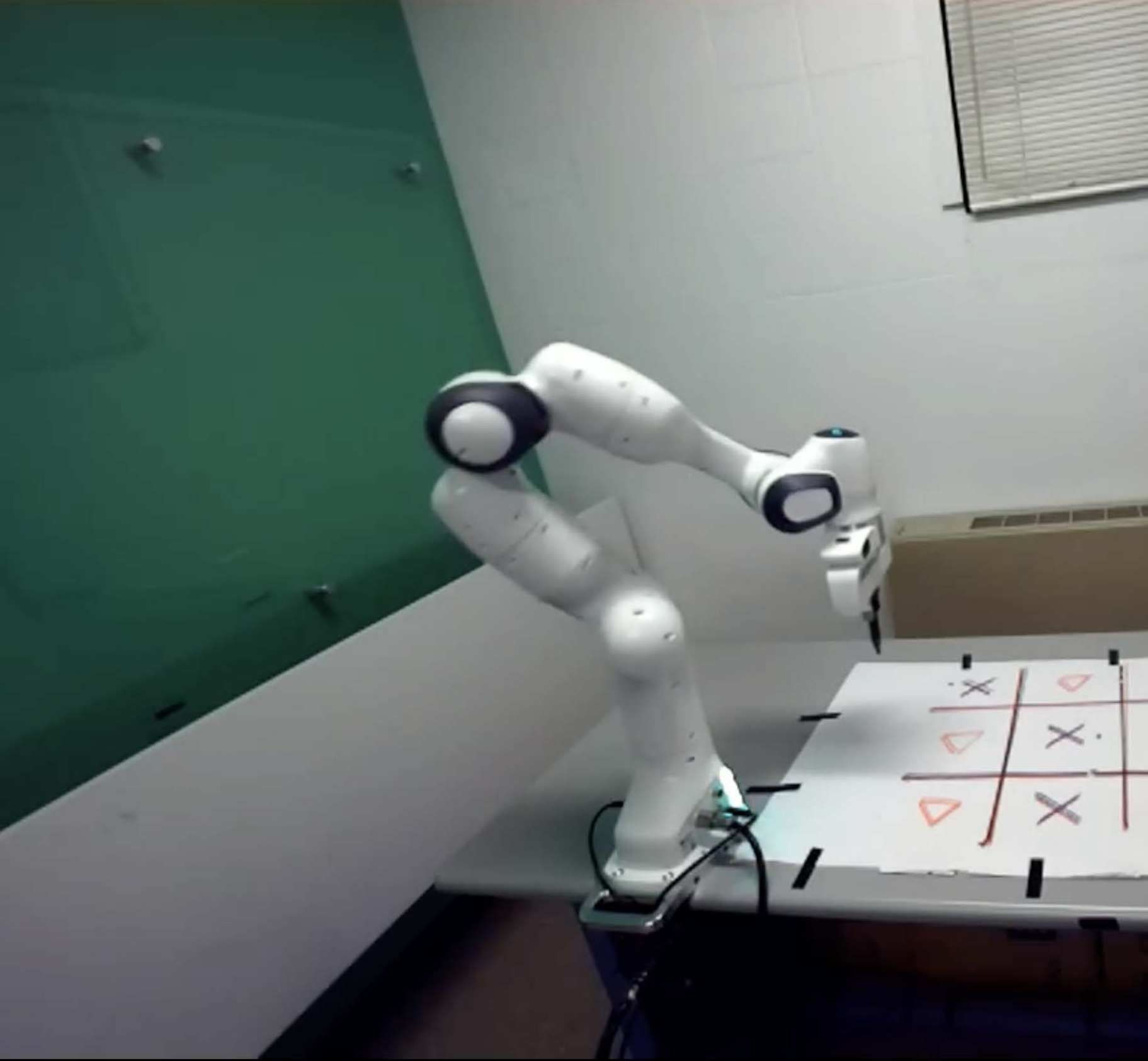}
      \end{minipage}\\
    \caption{Dataset samples: First row captured by camera one, second row by camera two as depicted in Figure~\ref{fig:Labsettig}.}
    
    \label{fig:pic2-Dataset}
\end{figure}

Each recorded video has $T=481$ frames captured over a $16$-second duration, with an original image size of $H=1280$ by $W=720$. To ensure the diversity of our dataset, we gathered $50$ samples for each of the nine action classes.  Figure~\ref{fig:pic2-Dataset} shows some samples of the actions. We run experiments using a dataset consisting of $100$ annotated frames.

\subsection{Training Setup} 
The network was trained for up to $10^5$ iterations with a batch size of $8$, and early stopping was utilized.
To select the optimizer and learning rate, we conducted a grid search involving learning rates from the set $\{0.001, 0.01, 0.1\}$ and considered three optimization algorithms: SGD, RMSprop, and Adam. The categorical cross-entropy loss function along with the Adam optimizer \cite{kingma2014adam} were configured with a learning rate of $0.001$.
The split percentages for all experiments are consistently set to 80$\%$, 10$\%$, and 10$\%$ for the training, validation, and the testing set, respectively.

\subsection{Result Analysis}
Our proposed model achieved high accuracy, with only a few misclassifications observed, between labels 8 and 5, as well as between labels 1 and 4. The proximity of these positions on the Tic-Tac-Toe board contributed to the confusion. In roughly 50 to 60 epochs, our model attains a testing accuracy of approximately 98$\%$, with a validation accuracy of around 97.7$\%$.
To assess the classification model's performance, we employ two distinct models, transformer and Rocket, for classifying robot arm movements. We then compare their results with our proposed method, particularly evaluating their performance in a noisy environment. 

\subsubsection{Baseline Models}
We utilized a transformer model including 4 attention heads, a point-wise feedforward network with a depth of 128, and 4 transformer encoder blocks, complemented by dense layers. The activation function used in the dense layers was ReLU. In addition, we incorporated 1-D global average pooling to compress the output generated by the transformer encoder module. 
Within approximately 80-90 epochs, our model achieves a testing accuracy of around 94.2$\%$, a validation accuracy of approximately 93.7$\%$. The Rocket model achieved a testing accuracy of around 95.6$\%$, a validation accuracy of approximately 93.8$\%$.

\subsubsection{Evaluation in Noisy Environments}
In this section, we perform a comparative analysis of three classification models by applying cut-out, salt and pepper, and Gaussian noise to time series data that represent the robot arm's trajectory. To understand the influence of noise, we designed and executed three experiments: the first employs noise augmentation only during training; the second tests the models on noisy data after training them on noise-free data; and the third both trains and tests the models on noisy data.
The study compares these experiments with models trained without any noise. The main aim is to assess the model's ability to generalize to noisy data, simulating situations like camera malfunctions where frames are missing, resembling cut-out noise.
Over the course of these experiments, we systematically increase the level of noise from 10$\%$ to 50$\%$, enabling us to explore the model's resilience to escalating levels of noise and its subsequent ability to maintain accuracy and robustness.
In the following, we provide detailed descriptions of three distinct types of noise applied to the dataset.

\textbf{Cut-out Noise:} In this approach, random rectangular regions within the time series are masked by setting their pixel values to zero. This technique encourages the model to understand context and adapt to missing or distorted information. 
The region size is 10, and masking is through a randomized probability selection process.

\textbf{Salt and Pepper:} Adding salt and pepper noise to time series data means inserting random spikes (salt) or drops (pepper) to mimic extreme values, making the data more resilient and ready for unexpected changes. We use parameters like density (how many data points get noise) and intensity (how strong the noise is). Density ranges from 10 to 50 in five steps, and intensity depends on the data's minimum (for pepper) or maximum (for salt) values. We decide to apply salt or pepper noise randomly to each data point based on probability.

\textbf{Gaussian Noise:} The critical hyperparameter in applying Gaussian noise is the noise level or standard deviation, which we have explored across a range of values, from 0.1$\%$ to 0.5$\%$.

\begin{figure}[H]
    \begin{center}
         \includegraphics[scale=0.32]{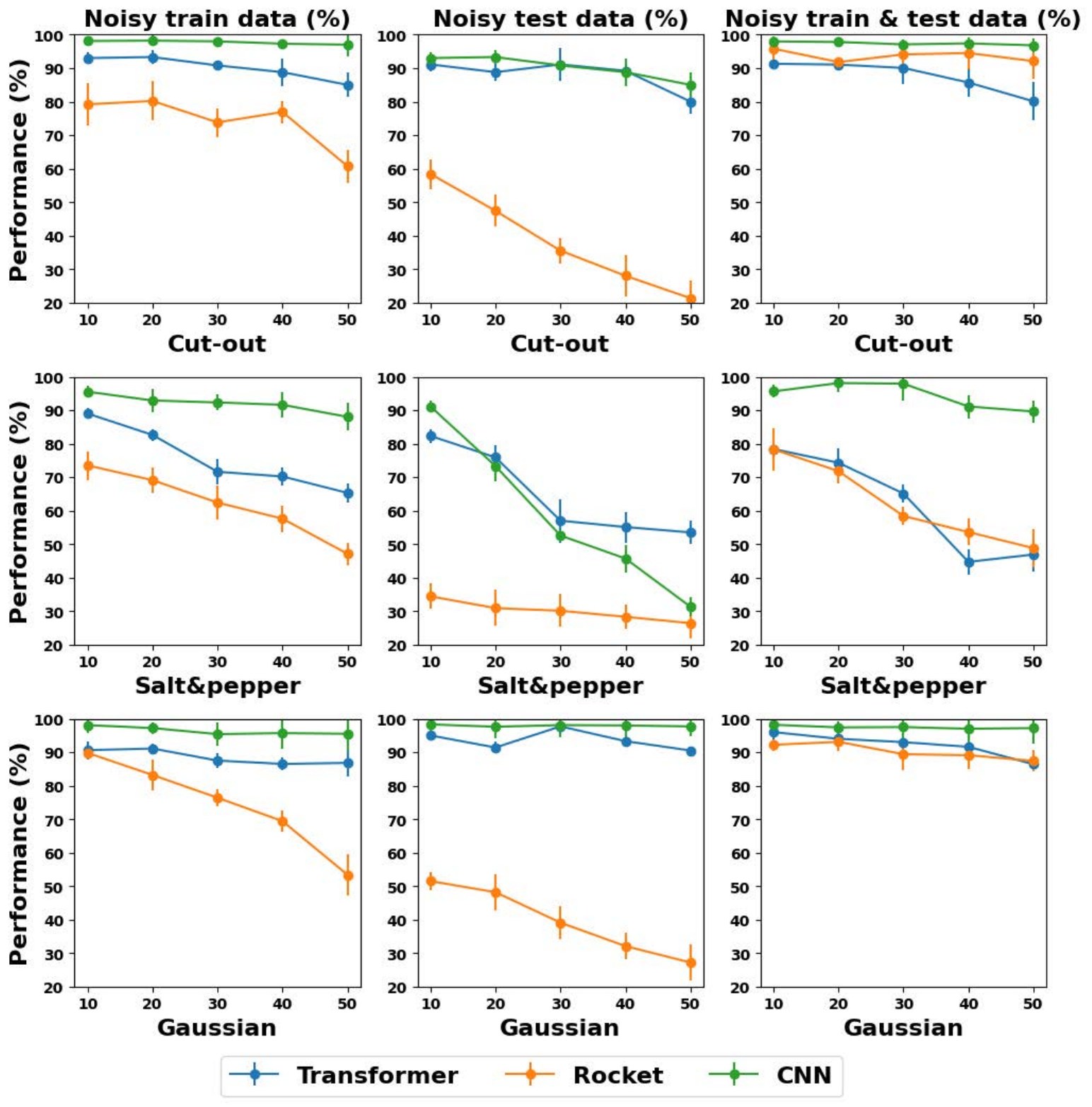}
    \end{center}
    \vspace{-0.3cm} 
    \caption{Comparative analysis of transformer, Rocket, and CNN models in diverse noise scenarios during training, testing, and combined, in percentage. Top row: Cut-out noise; middle row: Salt and Pepper noise; bottom row: Gaussian noise.}  
    \label{fig:subplot}
\end{figure}

Average and standard deviation (STD) of accuracy for transformer, Rocket and CNN models are presented in Tables~\ref{tab:transformer-comparison},~\ref{tab:rocket-comparison} and~\ref{tab:cnn-comparison} after 5-fold cross-validation.
Additionally, Figure~\ref{fig:subplot} provides a comprehensive analysis of how each type of noise impacts the performance of the three models, facilitating their comparison.
In general, these models may display sensitivity to data noise, with variations depending on their architectures and the data's characteristics. Our study, conducted within the context of time series data and our specific application, illustrates how the sensitivity of these models to noise is influenced by their architectures and the unique features of the data. 
In summary, the CNN model emerges as the most promising performer, demonstrating exceptional capability in excelling even in noisy environments. On the other hand, Rocket exhibits the lowest level of effectiveness, with a notable drop in accuracy when exposed to noise. However, when noise is introduced into both training and test datasets, the Rocket model's performance remains consistent with that in a noise-free environment. It is worth noting that among all types of noise, salt and pepper noise has the most detrimental impact on accuracy across all models.

\begin{table}[t!]
\footnotesize
    \centering
    \caption{Performance of the \textbf{transformer-based model} with noisy data. Noise and accuracy results are in $\%$.}
\begin{adjustbox}{width=0.44\textwidth}    
\begin{tabular}{cccccc}
\hline
\footnotesize{Noise}           & Noise type~      & Train ~&  Test~& Train \& Test \\ \hline
\multirow{3}{*}{10} & Cut-out         &  93.0$\pm$1.6 & 91.1$\pm$1.9&91.3$\pm$1.1 \\  
                   &Salt \& Pepper&89.0$\pm$1.6&82.3$\pm$2.1 &78.4$\pm$2.9\\ 
                   & Gaussian &90.6$\pm$2.7&95.0$\pm$1.1&96.0$\pm$2.1\\ 
                 \hline
\multirow{3}{*}{20} & Cut-out         &93.3$\pm$2.1 &88.8$\pm$2.7&91.0$\pm$1.7\\ 
                   & Salt \& Pepper &82.6$\pm$1.8& 75.9$\pm$3.5&74.3$\pm$4.4\\ 
                   & Gaussian &91.1$\pm$1.6&91.4$\pm$1.7&94.0$\pm$1.7\\ 
                    \hline
\multirow{3}{*}{30} & Cut-out         &90.8$\pm$1.5&91.1$\pm$4.8&90.6$\pm$4.9\\ 
                   & Salt \& Pepper &71.6$\pm$3.9 &57.0$\pm$6.5&65.1$\pm$2.7\\ 
                   & Gaussian &87.5$\pm$2.2&97.7$\pm$2.6&93.0$\pm$2.3\\ 
                \hline
\multirow{3}{*}{40} & Cut-out         &88.8$\pm$4.1&89.2$\pm$2.1&85.7$\pm$4.3\\  
                   & Salt \& Pepper &70.2$\pm$2.6&55.1$\pm$4.6&44.7$\pm$3.9\\ 
                   & Gaussian &86.5$\pm$1.8&93.3$\pm$1.6&91.6$\pm$3.8\\ 
                    \hline
\multirow{3}{*}{50} & Cut-out         &85.0$\pm$3.6&80.0$\pm$3.6&80.1$\pm$5.8\\  
                   & Salt \& Pepper &65.3$\pm$2.9 &53.5$\pm$3.6&46.9$\pm$5.1\\ 
                   & Gaussian &86.8$\pm$4.0&90.5$\pm$1.1&86.4$\pm$1.5\\ 
                    \hline               
\end{tabular}
\end{adjustbox}
    \label{tab:transformer-comparison}
\end{table}
\begin{table}[t!]
\footnotesize
    \centering
    \caption{Performance of the \textbf{Rocket model} with noisy data. Noise and accuracy results are in $\%$.}
\begin{adjustbox}{width=0.44\textwidth}    
\begin{tabular}{cccccc}
\hline
\footnotesize{Noise}           & Noise type      & Train&  Test &  Train \& Test\\ \hline
\multirow{3}{*}{10} & Cut-out& 79.2$\pm$6.2 & 58.4$\pm$4.4&95.8$\pm$3.7 \\  
                   &Salt \& Pepper&73.5$\pm$4.3&34.4$\pm$3.8 &78.3$\pm$6.3\\ 
                   & Gaussian &89.8$\pm$1.4& 51.5$\pm$2.7&92.2$\pm$1.7\\ 
                 \hline
\multirow{3}{*}{20} & Cut-out         &80.2$\pm$5.9&47.4$\pm$4.8&91.8$\pm$1.4\\ 
                   & Salt \& Pepper &69.1$\pm$3.8& 30.9$\pm$5.4&71.8$\pm$3.7\\ 
                   & Gaussian &83.2$\pm$4.6&48.2$\pm$5.3&93.1$\pm$2.6\\ 
                    \hline
\multirow{3}{*}{30} & Cut-out         &73.8$\pm$4.3&35.5$\pm$3.7&94.1$\pm$4.6\\ 
                   & Salt \& Pepper &62.4$\pm$5.2 &30.1$\pm$4.9&58.4$\pm$2.7\\ 
                   & Gaussian &76.4$\pm$2.4&39.1$\pm$4.8 &89.4$\pm$4.9\\ 
                \hline
\multirow{3}{*}{40} & Cut-out         &76.9$\pm$3.3&28.0$\pm$6.3&94.5$\pm$4.8\\  
                   & Salt \& Pepper &57.6$\pm$3.9&28.3$\pm$3.7&53.6$\pm$4.0\\ 
                   & Gaussian &69.4$\pm$3.3&32.7$\pm$3.9&89.1$\pm$4.1\\ 
                    \hline
\multirow{3}{*}{50} & Cut-out         &60.7$\pm$4.8&21.3$\pm$5.2&92.1$\pm$5.3\\  
                   & Salt \& Pepper &47.1$\pm$3.4 &26.4$\pm$4.5&48.8$\pm$5.7\\ 
                   & Gaussian &53.4$\pm$6.1&27.2$\pm$5.5&87.4$\pm$3.2\\ 
                    \hline                  
\end{tabular}
\end{adjustbox}
    \label{tab:rocket-comparison}
\end{table}

\section{Conclusion}
This study focuses on vision-based robot action recognition in noisy environments, where our model performs real-time, accurate 2-D pose estimation of a robot arm from video inputs. In contrast to studies focused on ideal environments, our research evaluates the performance of state-of-the-art neural networks like transformer and Rocket in noisy environments. Our findings show robustness of the model in robot arm action recognition in various noisy environments.

\begin{table}[t!]
\footnotesize
    \centering
    \caption{Performance of the \textbf{CNN model} with noisy data. Noise and accuracy results are in $\%$.}
\begin{adjustbox}{width=0.44\textwidth}    
\begin{tabular}{cccccc}
\hline
\footnotesize{Noise}           & Noise type      &  Train&  Test &  Train \& Test \\ \hline
\multirow{3}{*}{10} & Cut-out& 98.1$\pm$1.2&98.0$\pm$1.2&98.0$\pm$1.6 \\  
                   &Salt \& Pepper&95.5$\pm$1.8&91.1$\pm$1.8 &95.6$\pm$1.9\\ 
                   & Gaussian &98.1$\pm$2.4&98.3$\pm$2.2&98.2$\pm$3.1\\ 
                 \hline
\multirow{3}{*}{20} & Cut-out         &98.2$\pm$1.7&97.1$\pm$2.6&97.8$\pm$1.5\\ 
                   & Salt \& Pepper &92.8$\pm$3.4& 73.3$\pm$4.5&98.1$\pm$2.7\\ 
                   & Gaussian &97.2$\pm$1.7& 97.6$\pm$3.5 & 97.4$\pm$1.9\\ 
                    \hline
\multirow{3}{*}{30} & Cut-out         &98.0$\pm$1.1&98.2$\pm$1.7&97.1$\pm$1.4\\ 
                   & Salt \& Pepper &92.3$\pm$2.4 &52.5$\pm$1.7&97.9$\pm$4.9\\ 
                   & Gaussian &95.4$\pm$3.6&98.1$\pm$3.6 &97.5$\pm$3.8\\ 
                \hline
\multirow{3}{*}{40} & Cut-out         &97.3$\pm$1.3&97.0$\pm$1.5&97.4$\pm$1.9\\  
                   & Salt \& Pepper &91.6$\pm$3.9 &45.7$\pm$4.1&91.1$\pm$3.5\\ 
                   & Gaussian &95.7$\pm$4.8&98.0$\pm$4.7&97.0$\pm$4.1\\ 
                    \hline
\multirow{3}{*}{50} & Cut-out         &97.0$\pm$3.4&97.7$\pm$2.9&96.8$\pm$2.2\\  
                   & Salt \& Pepper &88.0$\pm$4.1&31.3$\pm$2.9&89.6$\pm$3.3\\ 
                   & Gaussian &95.5$\pm$5.2&97.7$\pm$3.0&97.2$\pm$4.6\\ 
                    \hline                  
\end{tabular}
\end{adjustbox}
    \label{tab:cnn-comparison}
\end{table}



\bibliographystyle{IEEEbib}
\ninept
\bibliography{strings,refs}

\end{document}